\documentclass[sigconf, nonacm, 10pt]{acmart} 
\AtBeginDocument{%
  }

\newcommand{\method}{\texttt{Sentry\-Fuse}}
\newcommand{\gate}{\texttt{Sentry\-Gate}}
\newcommand{\attn}{\texttt{Sentry\-Attend}}

\newcommand{\best}[1]{\textcolor{LinkColor}{\textbf{#1}}}   
\newcommand{\secondbest}[1]{\underline{#1}}                  

\newcommand{\localbest}[1]{\textcolor{LinkColor}{#1}}

\usepackage{xcolor}
\definecolor{LinkColor}{RGB}{153,51,51}

\usepackage[utf8]{inputenc}
\usepackage[T1]{fontenc}
\usepackage{microtype}
\usepackage{nicefrac}

\usepackage{amsmath}
\usepackage{amsfonts}
\usepackage{mathtools}
\usepackage{amsthm}
\usepackage{mathrsfs}

\usepackage{graphicx}
\usepackage{subcaption}
\usepackage{caption}
\usepackage{wrapfig}

\usepackage{booktabs}
\usepackage{tabularx}
\usepackage{multirow}
\usepackage{colortbl}
\usepackage[table,xcdraw]{xcolor}

\usepackage{hyperref}

\usepackage[ruled,vlined,linesnumbered]{algorithm2e}
\SetKwInOut{KwIn}{Input}
\SetKwInOut{KwOut}{Output}

\usepackage{pifont}

\usepackage{lipsum}
\usepackage{gensymb}
\usepackage{enumitem}
\usepackage{xspace}
\usepackage{ulem}
\useunder{\uline}{\ul}{}
\usepackage{capt-of}
\usepackage[capitalize,noabbrev]{cleveref}
\usepackage{booktabs}
\usepackage{multirow}
\usepackage{graphicx}
\usepackage{xcolor}
\usepackage{makecell}

\begin{document}
\renewcommand\footnotetextcopyrightpermission[1]{} 
\settopmatter{printacmref=false}                   

\title{Modality-Aware Zero-Shot Pruning and Sparse Attention for Efficient Multimodal Edge Inference}

\author{Yueyuan Sui}
\authornote{Both authors contributed equally to this research.}
\email{yueyuansui@u.northwestern.edu}
\affiliation{%
  \institution{Northwestern University}
  \city{Evanston}
  \state{Illinois}
  \country{USA}
}

\author{Payal Mohapatra}
\authornotemark[1]
\email{payal.mohapatra@u.northwestern.edu}
\affiliation{%
  \institution{Northwestern University}
  \city{Evanston}
  \state{Illinois}
  \country{USA}
}

\author{Do\u{g}a\c{c} Eldenk}
\affiliation{%
  \institution{Northwestern University}
  \city{Evanston}
  \state{Illinois}
  \country{USA}
}

\author{Haodong Yang}
\affiliation{%
  \institution{Northwestern University}
  \city{Evanston}
  \state{Illinois}
  \country{USA}
}

\author{Yiting Zhang}
\affiliation{%
  \institution{Northwestern University}
  \city{Evanston}
  \state{Illinois}
  \country{USA}
}

\author{Haoyan Zhang}
\affiliation{%
  \institution{Northwestern University}
  \city{Evanston}
  \state{Illinois}
  \country{USA}
}

\author{Qi Zhu}
\affiliation{%
  \institution{Northwestern University}
  \city{Evanston}
  \state{Illinois}
  \country{USA}
}

\author{Stephen Xia}
\affiliation{%
  \institution{Northwestern University}
  \city{Evanston}
  \state{Illinois}
  \country{USA}
}

\renewcommand{\shortauthors}{Trovato et al.}

\begin{abstract}

Edge devices increasingly run multimodal sensing pipelines that must remain accurate despite fluctuating power budgets and unpredictable sensor dropout. Existing pruning methods fail under these conditions: they generally require fine-tuning after compression, consuming over $10\times$ the deployment energy, and they assign static importance scores that are blind to which sensors are present. We present the \method{} framework, which addresses both challenges jointly through two key components. First, \gate{} learns modality-conditioned importance scores during training via first-order saliency supervision and then prunes attention heads and feed-forward channels at deployment without fine-tuning. Second, \attn{} replaces dense self-attention, a key bottleneck in contemporary multimodal architectures, with sparse grouped-query attention, yielding a net 15\% reduction in GFLOPs across three different multimodal architectures. Across three applications and multimodal backbones, \gate{} achieves a 12.7\% average accuracy improvement over the strongest pruning baseline, and upto to 18\% under modality dropout conditions. Together, \method{} reduces memory by 28.2\% and lowers latency by up to $1.63\times$ without further fine-tuning, establishing modality-aware zero-shot compression as a practical path to multimodal intelligence on heterogeneous edge hardware.

\end{abstract}

\ccsdesc[500]{Computing methodologies~Machine learning}
\ccsdesc[500]{Computer systems organization~Embedded and cyber-physical systems}



\keywords{edge inference, mobile sensing, multimodal time-series learning, model compression, structured pruning, sensor dropout, 
energy-constrained deployment, efficient attention}

\maketitle

\vspace{-10pt}
\section{Introduction}\label{sec:intro}

Edge-based continuous sensing systems are now ubiquitous due to rapid advances in IoT frameworks and sensing instrumentation~\cite{altun2010comparative, goldberger2000physiobank, ragab2023adatime}. Devices such as smartwatches incorporate numerous sensors that capture heterogeneous physiological and behavioral signals, such as heart rate, respiration, motion, and temperature, often at different sampling rates and possessing inherently diverse characteristics. Leveraging these complementary signals enables a wide range of applications such as activity recognition, stress monitoring, sleep quality estimation, and more~\cite{ragab2023adatime, fu2024remote, jones2025beyond, mohapatra2024wearable}. Consequently, recent research has increasingly focused on multimodal time-series learning frameworks that explicitly model both intra-modal temporal dynamics and inter-modal interactions~\cite{liang2021multibench, middlehurst2024bake}. These recent approaches go beyond heuristic sensor fusion methods and the direct application of pairwise cross-modal learning paradigms, which can become combinatorially expensive as the number of sensing modalities grows, especially beyond four modalities. Most smart devices today easily contain about a dozen sensors~\cite{masoumian2023smartwatches, schmidt2018introducing}. 

While these frameworks improve predictive performance, deploying them on mobile and edge platforms remains challenging. Real-world sensing systems operate under strict constraints on compute capability, memory bandwidth, and energy consumption. Mostcurrent multimodal models for temporal data leverage computationally-expensive transformer-based backbones. Moreover, sensing environments are inherently unreliable, often leading to individual sensor failures during deployment due to hardware faults, power limitations, or communication failures. Thus, efficient multimodal models must operate under incomplete observations.

Existing solutions try to address these challenges through model pruning and compression, particularly structured pruning techniques developed for transformer architectures~\cite{molchanov2016pruning, han2015learning, li2017pruning}. These approaches have been effective in reducing the computational cost of large models, ranging from large language models to time-series transformers~\cite{frantar2023sparsegpt, ma2023llm}. However, two critical considerations for practical multimodal deployment remain underexplored. First, many pruning methods require fine-tuning after pruning~\cite{molchanov2016pruning, he2017channel}, which is often infeasible directly on resource-limited devices. Second, existing strategies typically assume the presence of all modalities during inference, ignoring the occurrence of missing sensor streams~\cite{ma2022multimodal, ma2021smil}. These limitations expose three key challenges for deploying multimodal time-series architectures in practical edge sensing environments:

\textbf{Challenge 1: Limited and variable compute resources.} Mobile and edge devices operate under strict power and memory constraints, which restrict model size, complexity, and on-device trainability. 

\textbf{Challenge 2: Missing modalities.} Sensor failures or communication dropouts frequently lead to incomplete modality sets during inference~\cite{ma2022multimodal, ma2021smil}. Any solution tailored for mobile and edge scenarios must be capable of dynamically adapting to whichever sensor streams are present.

\textbf{Challenge 3: Lack of modality-awareness in pruning.} Existing pruning techniques~\cite{molchanov2016pruning, wang2021spatten} typically treat structural components independently of the sensing modalities they process, ignoring the cross-modal dependencies that drive representational synergy in multimodal architectures~\cite{liang2021multibench}. When applied naively, this can degrade cross-modal representations during compression. As we show in Figure~\ref{fig:attn_triplet}, this failure compounds severely under modality missingness.

These challenges also reveal two key opportunities for improving the efficiency and pragmatism of current multimodal time-series models.

\textbf{Opportunity 1: Efficient attention mechanisms.} Most multimodal time-series architectures use a transformer backbone, and self-attention is one of the key computational bottlenecks (Section~\ref{sec:prelims}). Previous works have also noted its redundancy, especially in attention heads. This calls for more drop-in replacement attention variants for such multimodal time-series models to reduce the number of operations and to find ways to structurally prune attention heads to reduce memory footprint.

\textbf{Opportunity 2: Modality-aware attention-head pruning.} Incorporating modality awareness into the saliency scoring mechanism (assigning importance to model parameters which forms the basis of structural pruning~\cite{molchanov2016pruning}) allows pruning strategies to preserve structurally important cross-modal components while enabling zero-shot compression without fine-tuning. This expands the utility of these multimodal time-series models for deployment on devices with variable power constraints and limited sensing modalities, as they can be trained once and then compressed effectively based on device requirements without additional fine-tuning.

Leveraging these insights, we propose \method{}, a robust framework for task-driven multimodal time-series analysis designed for mobile and edge deployment. \method{} adopts a two-fold strategy.

First, \method{} introduces a modality-aware pruning paradigm through \gate{}, a lightweight gating mechanism that learns adaptive importance scores for structural components conditioned on the active modality set, without propagating gradients through the base model~\cite{molchanov2016pruning, selvaraju2017grad}. At deployment, \gate{} enables \method{} to dynamically adjust the number of active parameters according to available power or compute constraints of the edge device in a zero-shot manner, while remaining robust to arbitrary subsets of modalities.

Second, \method{} addresses a major computational bottleneck in transformer architectures~\cite{vaswani2017attention}—the multi-head attention (MHA) mechanism—through \attn{}, a Sparse Grouped-Query Attention mechanism. In contrast to standard MHA, where each attention head independently generates queries, keys, and values, \attn{} shares key-value projections across groups of heads~\cite{ainslie2023gqa}, exploiting the redundancy of attention patterns observed in temporal sequences~\cite{zhou2021informer} and enabling substantial efficiency gains with minimal performance degradation. We summarize our \textbf{key contributions} below:

1. We make the case for \textbf{zero-shot pruning of multimodal time-series models} as a practical deployment necessity.
Real-world edge sensing environments impose two simultaneous constraints, variable power budgets and unpredictable sensor availability, that existing pruning methods fail to address jointly. To address these challenges, we propose \method{}, a fine-tuning-free framework that enables modality-aware pruning (\gate{}) and efficient attention mechanisms to alleviate the self-attention bottleneck in multimodal models (\attn{}).


2. We propose \gate{}, a \textbf{modality-conditioned saliency gating mechanism} for structured pruning without fine-tuning.
Unlike modality-agnostic strategies that assign static importance scores, \gate{} learns scores conditioned on the active modality set, enabling on-the-fly compression that natively accommodates missing sensors and variable power constraints. \gate{} achieves an average of 12.7\% accuracy improvement over the strongest baselines across a wide range of settings (144), with gains reaching upto 18\% under modality dropout.

3. We propose \attn{}, an \textbf{efficient drop-in replacement for dense self-attention}, the key computational bottleneck in multimodal architectures. By combining grouped-query attention with sparse query selection, \attn{} reduces floating-point operations by an average of 15\% and upto 29\% while preserving predictive performance, improving inference throughput without retraining.

4. We carry out extensive evaluations demonstrating the \textbf{end-to-end practicality of \method{}} across diverse hardware platforms and applications. Experiments on embedded platforms and mobile devices (Jetson TX2, iPhone~13~Pro, Google Pixel~8) show that \method{} simultaneously improves accuracy, reduces memory, and lowers latency relative to state-of-the-art baselines. We further demonstrate its compatibility with post-training quantization.

\section{Related Works}\label{sec:related}

\noindent \textbf{Pruning Methods.} Network pruning has long been used to shrink model size and reduce computation by removing redundant parameters or structures~\cite{han2015learning,molchanov2016pruning,frankle2018lottery}. Early work largely focused on \emph{unstructured} magnitude pruning, zeroing individual weights based on their norms to obtain high compression ratios~\cite{han2015learning}, but such unstructured sparsity brings limited acceleration on general hardware. Subsequent methods therefore moved to \emph{structured} pruning of channels, kernels, blocks, or layers so that the resulting subnetworks can be executed efficiently~\cite{li2017pruning,tai2015convolutional,he2017channel}. For Transformers and LLMs, prior work prunes attention heads, FFN channels, or full layers using weight magnitude, gradient/Taylor saliency, Hessian-based approximations~\cite{molchanov2016pruning}, data-free criteria such as SynFlow~\cite{tanaka2020synflow}, or attention-based schemes like SpAttn that aggregate multi-layer attention maps into head-importance scores~\cite{wang2021spatten}, and combines these with block- or layer-wise structured compression plus post-training quantization to further reduce memory and FLOPs~\cite{frantar2023sparsegpt,ma2023llm}. However, most methods still learn a single pruned structure per model under a fixed input distribution, without modeling modality-dependent importance or platform-level power constraints; by contrast, we use first-order saliency only during training to supervise a lightweight Dynamic Task-Aware Gating module that enables modality-aware, budget-constrained structured pruning at deployment without extra fine-tuning.

\smallskip
\noindent\textbf{Mechanistic Interpretability.} Gradient-based methods have been foundational in mechanistic interpretability, enabling researchers to map functional pathways and causally link neural components to model predictions. Classic techniques such as vanilla gradients and saliency maps reveal how perturbations to inputs or activations impact outputs \cite{smoothgrad2017}, while Grad-CAM extends these concepts to CNNs by visualizing class-discriminative regions \cite{selvaraju2017grad}. Grad-SAM adapts gradient-based explanations for attention-based models \cite{gradsam2022}, and gradient-based pruning strategies leverage Taylor expansions to assess parameter saliency \cite{molchanov2017pruning}. Orthogonal to gradient-based approaches, linear and shapelet decomposition methods offer a complementary lens for interpretable time-series classification~\cite{pandey2026timesliver,wen2025shedding}. However, gradient computation is often infeasible on resource-constrained platforms, and in this work, we design dynamic gates that predict approximate importance without backpropagation, enabling zero-shot compression on edge devices. However, gradient computation is often infeasible on resource-constrained platforms. In this work, we design dynamic gates that predict approximate importance without backpropagation, enabling zero-shot compression on edge devices.

\smallskip
\noindent\textbf{Learning under Missing Modalities.} A fundamental challenge in multimodal systems is handling arbitrary modality dropout. Two paradigms dominate: (i) \textit{reconstruction methods} that impute missing modalities \cite{ma2021smil, ma2022multimodal, sun2025enhancing}, which become computationally prohibitive as $M$ increases, and (ii) \textit{transfer methods} that leverage cross-modal dependencies through MoE-based routing \cite{NEURIPS2024_b2f2af54, NEURIPS2024_7d62a85e, wudynamic}, modality binding \cite{girdhar2023imagebind, mohapatra24_interspeech, mohapatra-etal-2025-llms}, or shared embeddings \cite{wang2023multi}. Since arbitrary missingness is prevalent in practical deployments, in this work we pursue modality-aware joint optimization of task performance and model compression, contributing a pragmatic framework for efficient and robust multimodal time-series learning.

\section{Preliminary Studies}
\label{sec:prelims}

Deploying multimodal time-series models on edge platforms requires robustness to two practical realities: \textit{variable compute budgets} imposed by hardware constraints and \textit{arbitrary modality missingness} arising from sensor failures, power-gating, or intermittent connectivity. We conduct four targeted measurements using contemporary multimodal time-series models and standard pruning techniques on the WESAD stress detection dataset~\cite{schmidt2018introducing}, which contains ten modalities for a three-class classification task (dataset details are provided in Section~\ref{sec:exp}), to expose limitations of existing approaches and motivate our \method{} framework.

\begin{figure}[!htbp]
    \centering
    \includegraphics[width=0.9\linewidth]{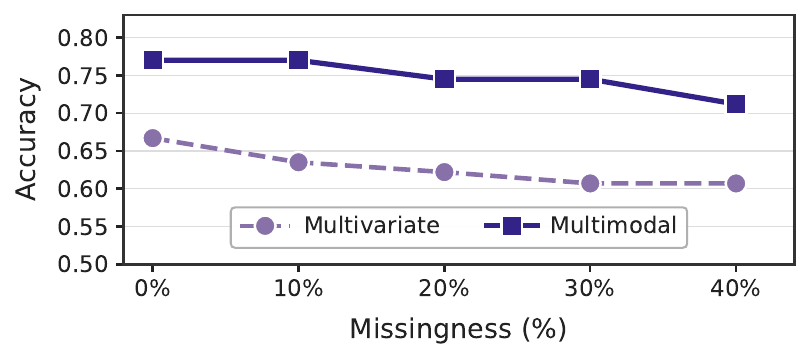}
    \vspace{-15pt}
    \caption{Multimodal~\cite{mohapatra2025maestro} vs. Multivariate~\cite{vaswani2017attention} modeling of heterogeneous sensor data under missingness.}
    \vspace{-10pt}
    \label{fig:mult_motivate}
\end{figure}

\paragraph{Observation 1: Multimodal Modeling of Heterogeneous Sensor Data Outperforms Multivariate Baselines, Especially Under Missingness.} Classical multivariate time-series models project all sensor channels into a shared latent space and model their joint dynamics via standard self-attention~\cite{vaswani2017attention,liu2023itransformer}. While generally effective, this design relies on uniform sampling rates across signals, and the absence of even a single modality can severely deteriorate the latent representation of that time instance. Hence, contemporary multimodal formulations and architectures aim to address this limitation and enhance the multimodal context of the heterogenous sensors by facilitating intra-modal and cross-modal attention strategies. Figure~\ref{fig:mult_motivate} illustrates the advantages of a multimodal architecture over a multivariate one, on a stress monitoring task: the multivariate model degrades by over 7\% as missingness increases from 0\% to 40\%, whereas the multimodal~\cite{mohapatra2025maestro} framework—which applies disentangled per-modality encoding followed by explicit cross-modal attention—sustains roughly 10\% higher accuracy throughout. The gap widens with increasing missingness, confirming that the multimodal architectural paradigm is a suitable path forward and that concrete strategies are needed to improve its pragmatism and efficiency for real-world deployment.

\begin{figure}[!htbp]
    \centering
    \includegraphics[width=\linewidth]{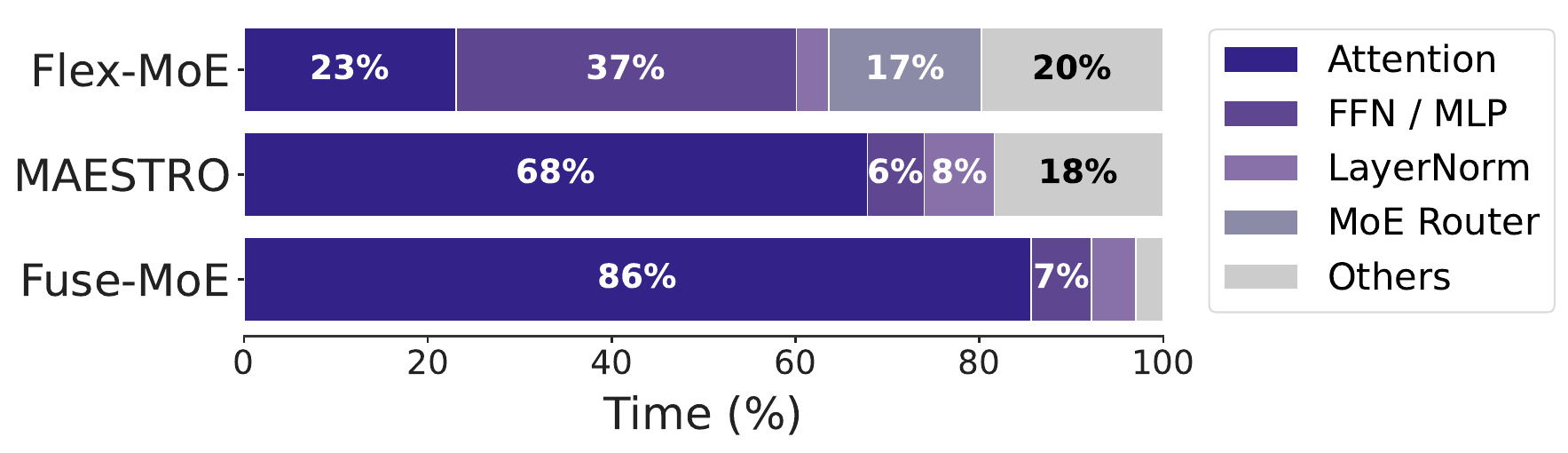}
    \vspace{-20pt}
    \caption{Time-spent profile during the Transformer forward pass in various multimodal architectures.}
    \label{fig:sa_motivate}
\end{figure}

\paragraph{Observation 2: Self-Attention is the Dominant Bottleneck in Multimodal Architectures.} Profiling different multi-modal backbones on a Jetson TX2 reveals that self-attention accounts for up to 83\% of inference latency (Figure~\ref{fig:sa_motivate}). This disproportionate cost stems from the quadratic complexity of causal attention over long time series, $\mathcal{O}(T^2D)$, where $T$ is the sequence length and $D$ is the hidden dimension. Any practical compression strategy must therefore target attention directly, rather than merely reducing parameter count, to yield meaningful latency reductions on edge hardware. 


\begin{figure}[!htbp]
    \centering
    \includegraphics[width=\linewidth]{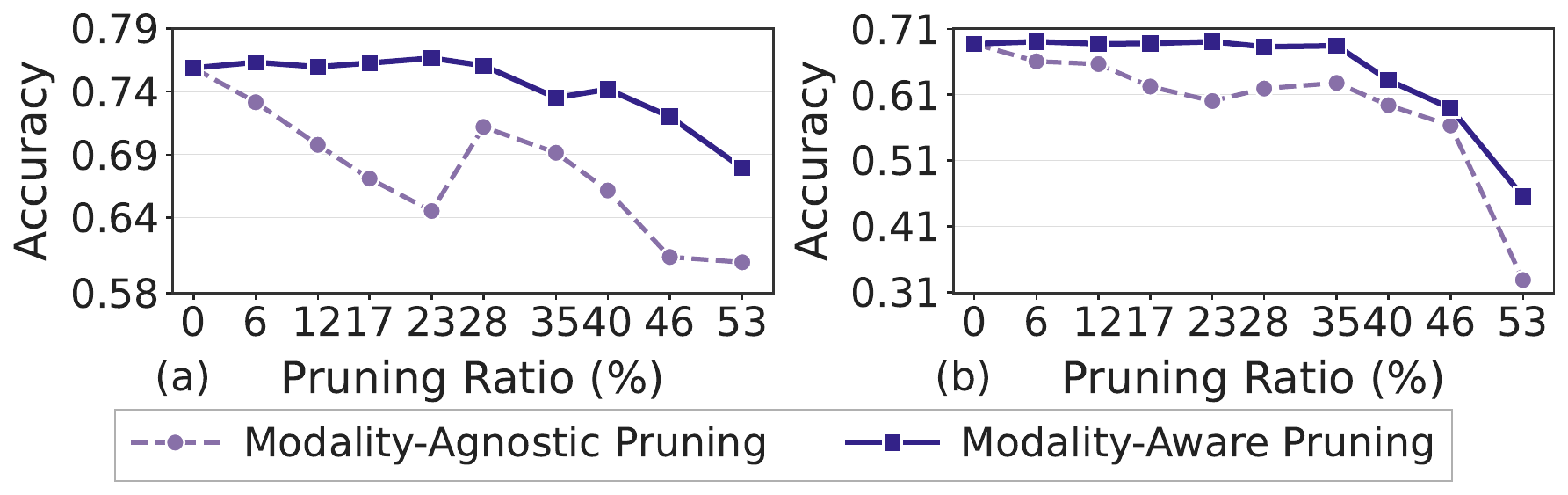}
    \vspace{-20pt}
    \caption{Performance of modality-aware pruning (\gate{}) and modality-agnostic pruning (SynFlow~\cite{tanaka2020synflow}) applied to transformer heads in a multimodal baseline (MAESTRO~\cite{mohapatra2025maestro}) (a) with all modalities present and (b) with 40\% modality dropout.}
    \label{fig:modality_prune}
    \vspace{-15pt}
\end{figure}

\paragraph{Observation 3: Modality-agnostic Pruning is Suboptimal, Especially Under Sensor Missingness.} Simple structured pruning methods, such as absolute magnitude-based scoring or advanced methods like SynFlow~\cite{tanaka2020synflow}, were designed for unimodal architectures under a fixed input distribution. Figure~\ref{fig:modality_prune} reveals their fundamental unsuitability for multimodal deployment. Even under complete modality observations, modality-agnostic pruning using SynFlow underperforms with respect to our proposed modality-aware \gate{} scoring mechanism, and the overall performance is further exacerbated when 40\% of modalities are missing. The root cause is that modality-agnostic criteria assign static importance scores to attention heads, ignoring the fact that the same head may be critical under one modality configuration and redundant under another. Hence, we need to support a saliency assignment strategy conditioned on the available modalities to advance the practical integration of multimodal time-series architectures.

\begin{figure}[!htbp]
    \centering
    \includegraphics[width=\linewidth]{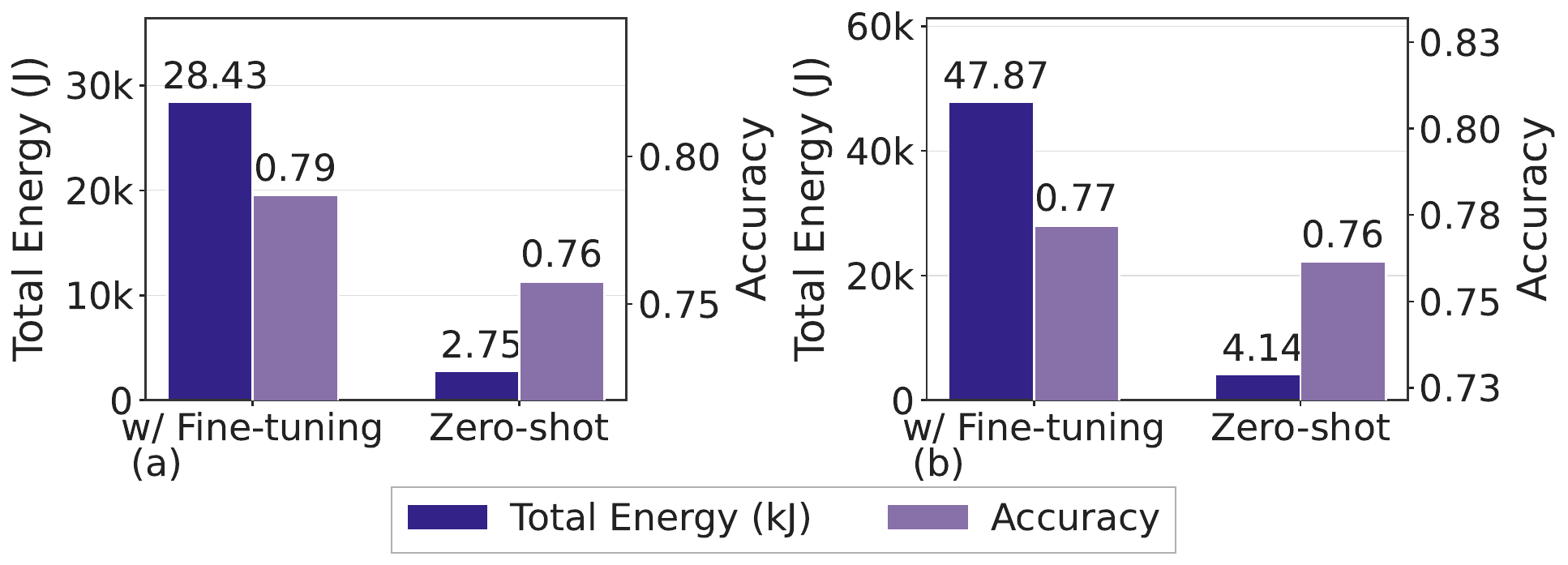}
    \vspace{-20pt}
    \caption{Total energy utilization in deploying a pruned model (23\%) with a fine-tuning (3 epochs) and in zero-shot across (a) Jetson TX2 and (b) an NVIDIA L40 GPU.}
    \vspace{-20pt}
    \label{fig:zeroshot_motivate}
\end{figure}

\paragraph{Observation 4: Zero-Shot Pruning is Essential for Practical Edge Deployment.} 
Standard pruning pipelines require a fine-tuning stage after compression to recover accuracy, which assumes the ability to perform backpropagation on the target device. This assumption fails on power-constrained edge platforms, where retaining intermediate activations for gradient computation can exhaust the available RAM at typical multimodal model scales, and sustained computation draws prohibitive cumulative energy. A pilot analysis in Figure~\ref{fig:zeroshot_motivate} demonstrates over ten times higher total energy utilization during fine-tuning of the pruned model compared to its zero-shot counterpart, while incurring only a marginal accuracy reduction ($\leq 0.03$). Our proposed \gate{} eliminates this overhead entirely by learning modality-conditioned importance scores during training and querying them once offline, enabling the once-trained, deploy-many-times paradigm demanded by power- and memory-constrained edge devices while remaining modality-aware.

\smallskip
\noindent\textbf{Implications.} The four observations above collectively inspired the design of \method{}. First, the backbone must disentangle modalities to remain accurate under arbitrary sensor dropout. Second, pruning must be conditioned on the active modality set, since static importance scores are insufficient—addressed by \gate{}. Third, compression must specifically target self-attention to address the dominant latency bottleneck—addressed by \attn{}. Finally, the entire pipeline must operate zero-shot at deployment, eliminating the fine-tuning dependency that current methods impose. Our proposed \method{} framework addresses all four requirements within a unified training objective.

\section{Approach}\label{sec:approach}

\subsection{System Overview}

Encouraged by the success of recent multimodal time-series  frameworks~\cite{mohapatra2025maestro, xu2025lsm, yun2024flex}, we propose \method{}, a framework for efficient and robust multimodal learning under variable compute and modality constraints. \method{} comprises two core components. First, \method{} introduces a plug-in \gate{} module (Sec.~\ref{sec:dtag}) trained using a custom saliency alignment objective (Sec.~\ref{sec:alignment}) conditioned on the available modalities. \gate{} learns adaptive importance scores for structural units conditioned on the active modality set. By supervising \gate{} with first-order saliency signals and jointly optimizing task and alignment losses, \method{} enables modality-aware pruning, resource-adaptive compression, and conditioned zero-shot deployment under varying energy budgets and modality availability. Second, \method{} replaces the dense self-attention layers of the transformer-based multimodal backbone with \attn{} (Sec.~\ref{subsec:sgqa}), a grouped sparse query attention mechanism that significantly reduces the number of operations during a forward pass while maintaining performance, making multimodal models more suitable for edge deployment.

\begin{figure*}[t]
    \vspace{-6pt}
    \centering
    \includegraphics[width=\linewidth]{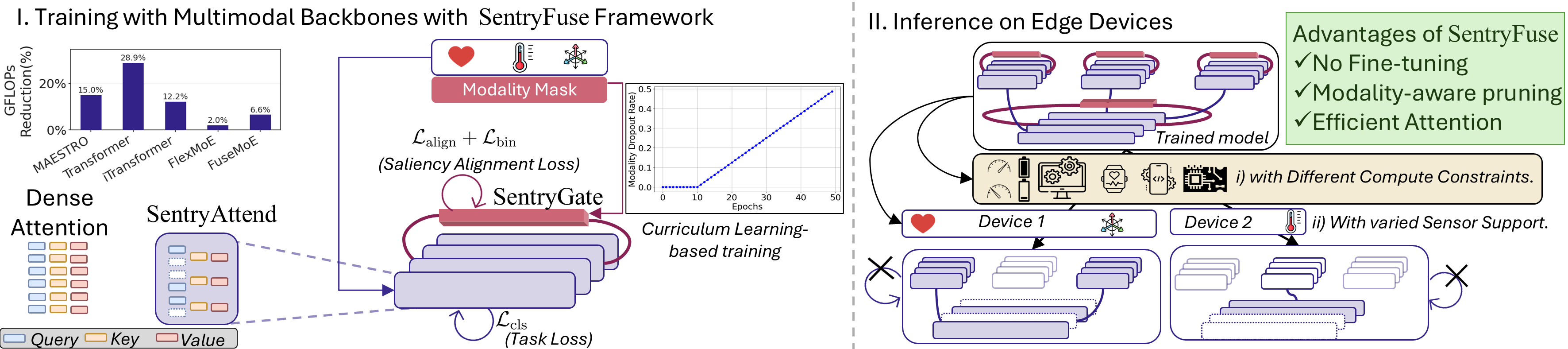}
    \vspace{-15pt}
    \captionsetup[]{font=small}
    \caption{Overview of the \method{} framework. I. During training (left), the multimodal backbone integrates \attn{}, replacing dense self-attention with sparse grouped-query attention for an average 13\% FLOP reduction. Simultaneously, \gate{} runs in observer mode, learning modality-conditioned structural importance under a curriculum of increasing modality dropout, guided by a saliency alignment loss. The backbone is optimized via the task loss $\mathcal{L}_{\text{cls}}$, while gate parameters receive gradients only from the alignment objective. II. At inference (right), \gate{} is queried once offline to produce a subnetwork given available modalities and a target compute budget, enabling a train-once, deploy-many paradigm that adapts to (i) varying compute constraints and (ii) arbitrary sensor availability across heterogeneous edge devices.}
    \vspace{-10pt}
    \label{fig:overall}
\end{figure*}

As illustrated in Fig.~\ref{fig:overall}, during training the \method{} framework is optimized using the supervised objective, while the \gate{} networks are trained using saliency and alignment objectives under a curriculum learning strategy that progressively exposes the model to missing modalities. During inference, based on the available modalities and the required compute budget, a compact subnetwork is realized on-the-fly. This enables a once-trained, deploy-many-times paradigm without any fine-tuning under different modality combinations on the deployed platform.

\subsection{\gate{}: Modality-aware Structural Gating}
\label{sec:dtag}

Multimodal backbone~\cite{mohapatra2025maestro, yun2024flex, NEURIPS2024_7d62a85e} components consist of dense structural units—multi-head attention and feed-forward channels, including those within Sparse-MoE blocks—that may be unnecessary under specific sensing configurations. To compress the model without fine-tuning, we aim to structurally eliminate unimportant components such as attention heads and feed-forward neurons in a manner conditioned on which modalities are present.

Designing such structural pruning requires addressing two key considerations. First, the importance of a structural unit should depend on which modalities are available at inference time: the same head or neuron may be crucial when a particular sensor is present but redundant when that modality is absent. Second, we need a resource-efficient strategy to infer structural importance at deployment, enabling the model to support different FLOPs or energy budgets on demand without repeated fine-tuning.

To address the first consideration, we expose the backbone to a rich space of modality configurations during training. Following recent multimodal strategies~\cite{mohapatra2025maestro,xu2025lsm}, we (i) encode missing modalities using symbolic tokens so the encoder can explicitly sense which channels are absent, and (ii) adopt a curriculum with progressively increasing modality dropout, allowing the model to stabilize under full-modality inputs before gradually adapting to severe missingness patterns.

An alternative for the second consideration would be to score structural units by their first-order sensitivity at inference time. In practice, this is incompatible with resource-constrained deployment: it requires a backward pass for each input—utilizing at least 10$\times$ the total energy to our zero-shot approach, as shown in Figure~\ref{fig:zeroshot_motivate}—as well as storing intermediate activations and sometimes accessing task labels that are unavailable during inference. Moreover, such scores are tied to individual samples rather than modality patterns, making them difficult to reuse across platforms and operating points.

To address this in a principled and deployment-friendly manner, we introduce \textbf{Modality-aware Structural Gating (\gate{})}, a lightweight gating module that predicts per-unit importance from the current modality mask and drives a fine-tuning-free pruning procedure. During training, first-order sensitivity serves only as a teacher signal to supervise \gate{}, which learns a compact, forward-only surrogate that can be queried cheaply at inference. Once trained, \gate{} outputs are evaluated offline under a target platform configuration and thresholded to select subnetworks meeting a desired FLOPs or energy budget. Changing budgets or platform masks requires no additional fine-tuning.

Formally, for the $i$-th structural unit in layer $\ell$ (an attention head 
or feed-forward channel, including those inside Sparse-MoE blocks), \gate{} 
maps the modality-availability mask $m \in \{0,1\}^M$ to a scalar 
importance score:
\begin{equation}
g_i^{(\ell)}(m) =
\sigma\Big(
\zeta_i^{(\ell)} +
\gamma^{(\ell)} f_{\phi_i}^{(\ell)}(m)
\Big),
\label{eq:dtag}
\end{equation}
where $\zeta_i^{(\ell)}$ is a learnable base importance, 
$f_{\phi_i}^{(\ell)}(\cdot)$ is a two-layer multi-layer perceptron (MLP) with a nonlinearity, 
and $\gamma^{(\ell)}$ is a layer-wise scaling factor. Compared to the backbone, \gate{} adds only a small parameter overhead, accounting for 9.10\%, 4.96\%, and 4.97\% of the total model parameters on WESAD, DaliaHAR, and DSADS, respectively, while its FLOPs overhead remains negligible.

During training, \gate{} operates in an \emph{observer} mode: gate outputs do not update the activations of the backbone, and the task loss $\mathcal{L}_{\text{cls}}$ does not flow through the pruning pathway. 
Gate parameters are optimized only via the saliency-alignment objective described in Section~\ref{sec:alignment}, which encourages $g_i^{(\ell)}(m)$ to approximate the first-order sensitivity of each unit under the current modality pattern. In this way, \gate{} learns a smooth, modality-conditioned importance function that captures how the relevance of each head or channel changes as sensors appear or disappear.

\subsection{Construction of Saliency Objective} \label{sec:alignment}

\gate{} aims to provide modality-aware importance scores for structural units, but these scores are not directly observed. As discussed in Section~\ref{sec:dtag}, we cannot estimate them at inference time 
by backpropagating gradients, since this would require storing the full computation graph, intermediate activations, and sometimes labels on resource-constrained platforms. Other inference-time proxies such as cumulative attention maps~\cite{wang2021spatten} aggregate attention weights across layers and heads, but we find them unstable under missing modalities and poorly correlated with downstream performance (see Sec.~\ref{sec:eval}). Instead, we use gradient-based signals only during training to construct a robust saliency target, and train \gate{} to approximate this target with a cheap forward computation usable at deployment.

To ensure that gate predictions reflect task relevance, we adopt a standard first-order saliency measure as the supervision signal. Let $\mathcal{L}_{\text{cls}}$ denote the classification loss and $x_i$ the 
activation associated with unit $i$. For a small perturbation $\delta x_i$, a first-order Taylor expansion yields,
\begin{equation}
\mathcal{L}_{\text{cls}}(x + \delta x) \approx 
\mathcal{L}_{\text{cls}}(x) +
\sum_i \frac{\partial \mathcal{L}_{\text{cls}}}{\partial x_i} \, \delta x_i.
\label{eq:taylor}
\end{equation}
Following common practice in gradient-based pruning and 
attribution~\cite{molchanov2016pruning,selvaraju2017grad}, we define the 
saliency of unit $i$ as
\begin{equation}
s_i = \big| x_i \odot \frac{\partial \mathcal{L}_{\text{cls}}}{\partial x_i} \big|,
\label{eq:saliency}
\end{equation}
which measures the magnitude of its contribution to changes in the loss. In practice, we aggregate $s_i$ over batch and time, and normalize it within 
each layer to obtain a target in $[0,1]$:
$\tilde{s}_i = \text{Norm}(s_i).$

Given the gate outputs $g_i \in [0,1]$ produced by \gate{} for the same units, we encourage them to match the normalized saliency targets via a mean-squared alignment loss:
\vspace{-10pt}
\begin{equation}
\mathcal{L}_{\text{align}} = 
\frac{1}{N} \sum_i \big( 
g_i - \tilde{s}_i
\big)^2,
\label{eq:align}
\end{equation}
where $N$ is the number of gated units in the layer. This loss encourages 
$g_i$ to preserve both the ranking and relative scale structure of 
$\tilde{s}_i$. To encourage discrete yet stable gating, we further 
introduce a mild binarization regularizer
\begin{equation}
\mathcal{L}_{\text{bin}} = \frac{1}{N} \sum_i g_i (1 - g_i),
\label{eq:binarize}
\end{equation}
which encourages gate outputs to move toward binary decisions while remaining fully differentiable during training.

The saliency alignment objective is complementary to $\mathcal{L}_{\text{cls}}$. Gate parameters are optimized using $\mathcal{L}_{\text{align}}$ and $\mathcal{L}_{\text{bin}}$, while backbone parameters are optimized primarily with respect to $\mathcal{L}_{\text{cls}}$, with gradient-based saliency treated as a detached target. In this way, \gate{} learns a modality-conditioned importance predictor during training and produces pruning scores at inference time using only a lightweight forward computation.



\subsection{\attn{}: Sparse Grouped-Query Attention}
\label{subsec:sgqa}

As we showed earlier in Figure~\ref{fig:sa_motivate}, self-attention is the key bottleneck across current multimodal architectures. We denote the multimodal time-series input as $\mathbf{X} \in \mathbb{R}^{M \times T}$, where $M$ is the number of modalities and $T$ is the number of time steps. A standard Transformer~\cite{vaswani2017attention} maps $\mathbf{X}$ to queries, keys, and values $\mathbf{Q}, \mathbf{K}, \mathbf{V} \in \mathbb{R}^{T \times F}$. For multi-head self-attention with $n$ heads, each head maintains its own $Q/K/V$ projection matrices, requiring approximately $3n \cdot M \cdot F$ parameters. In multimodal time-series settings, different attention heads often attend to highly correlated temporal patterns, making fully independent $K/V$ projections per head redundant.

To address this, we propose \attn{}, which adopts Grouped-Query Attention (GQA)~\cite{ainslie2023gqa}, partitioning the $n$ heads into $n_{\text{groups}}$ groups ($n_{\text{groups}} < n$). Each head retains its own query projection, while all heads within the same group share a common set of $K/V$ projections, reducing $K/V$ parameters from $2n \cdot M \cdot F$ to $2 n_{\text{groups}} \cdot M \cdot F$.

\begin{figure}[!htbp]
    \vspace{0pt}
    \centering
    \includegraphics[width=1.0\linewidth]{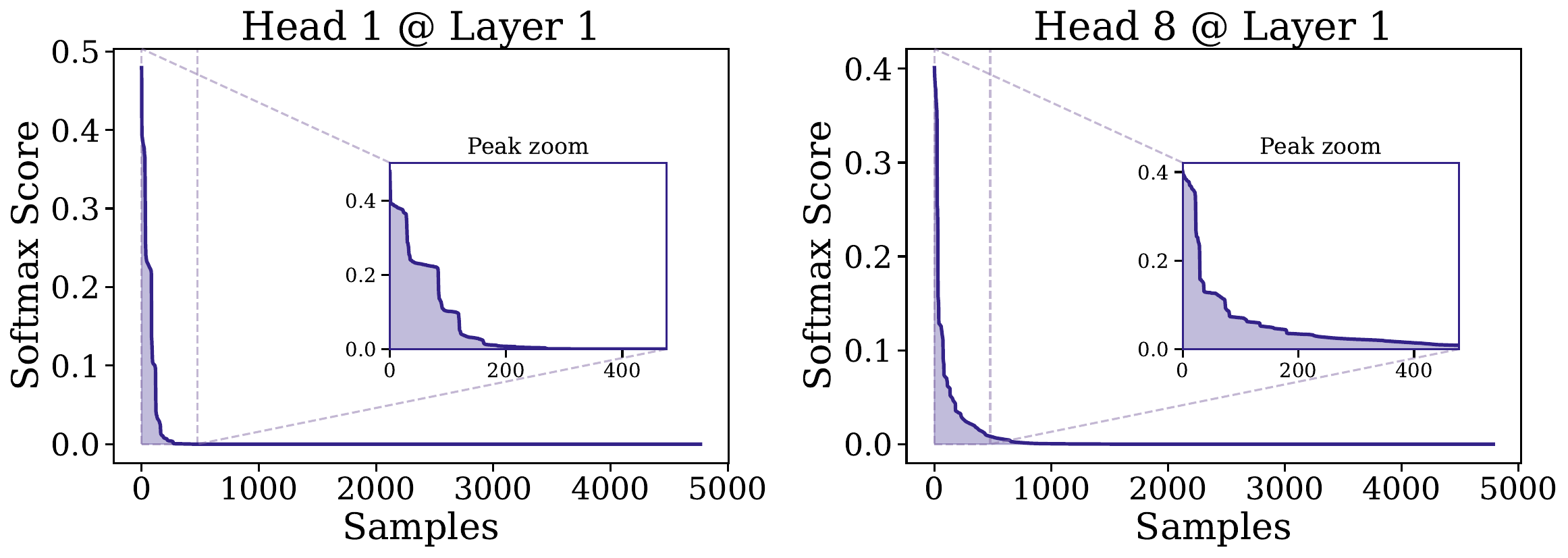}
    \vspace{-20pt}    
    \caption{Empirical long-tail self-attention patterns from two representative heads show that only a small fraction carry significant probability mass while the majority are nearly zero, motivating \attn{}.}
    \label{fig:tail}
\end{figure}

We leverage the observation from Figure~\ref{fig:tail}, where we take the softmax attention scores of a trained model, flatten them, and sort them in descending order. The results show that only a small fraction of positions account for the vast majority of the probability mass—the attention distribution remains strongly long-tailed, consistent with prior work~\cite{zhou2021informer,mohapatra2025maestro}. Motivated by this, \attn{} further introduces sparse groupe-query attention: within each head $h$ in group $g$, we compute a sparsity score for each query~\cite{zhou2021informer} and retain only the $U = c \cdot \lceil \log T \rceil$ queries with the highest sparsity scores, where $c$ is a constant controlling the sparsity level (typically $c=5$). The attention output is then computed as:
\begin{equation}
\vspace{-10pt}
\mathbf{A}_h = \text{Softmax}\left(\frac{\mathbf{Q}_h^{\text{top-}U} 
(\mathbf{K}_g)^\top}{\sqrt{d_k}}\right) \mathbf{V}_g,
\label{eq:sparse_gqa}
\end{equation}
where $\mathbf{Q}_h^{\text{top-}U} \in \mathbb{R}^{U \times d_k}$ contains 
only the top-$U$ selected queries, and $\mathbf{K}_g, \mathbf{V}_g$ are the 
shared key-value projections for group $g$. The remaining $T - U$ positions 
are filled with a context vector aggregated from $\mathbf{V}_g$, preserving 
output length while reducing per-head complexity from $\mathcal{O}(T^2)$ to 
$\mathcal{O}(T \log T)$.

\attn{} thus reduces GFLOPs along two independent dimensions: subsampling queries lower the temporal attention cost, and compressing $K/V$ projections reduces parameter and projection overhead. For example, with $n = 8$, $n_{\text{groups}} = 2$, $T = 128$, and $c = 5$, we have $U \approx 25$, so the attention computation scales from $\mathcal{O}(n T^2)$ to $\mathcal{O}(n U T)$, while the number of 
$K/V$ projection matrices is reduced from $2n$ to $2 n_{\text{groups}}$.

\subsection{Overall Optimization of \method{}}
\label{sec:overall_opt}

\textbf{Training procedure.} We jointly optimize the sparse multimodal encoder and \gate{} under a 
single objective. Each epoch proceeds with two coupled components. First, we update the backbone parameters to minimize $\mathcal{L}_{\text{cls}}$ under randomly sampled modality-availability patterns. Second, after a short warmup phase, we train \gate{} to align with first-order saliency while keeping it detached from the main task-loss masking path.

Concretely, let $m \in \{0,1\}^M$ denote the modality-availability mask. At epoch $t$, we draw $m$ from a curriculum-style schedule~\cite{mohapatra2025maestro} by sampling each modality as present with probability $1 - p_t$, where $p_t$ increases linearly from $0$ to $p_{\max}$ after a warmup period. For each mini-batch $(x,y)$:
\[
\hat{y} = f_\theta(x, m;\, \texttt{gates\_apply}=\texttt{False}), \qquad
\mathcal{L}_{\text{cls}} = \text{CE}(\hat{y}, y).
\]
During the first $T_{\text{warmup}}$ epochs, \gate{} parameters are frozen and we backpropagate only $\mathcal{L}_{\text{cls}}$. After warmup, we additionally invoke the saliency-alignment objective from 
Sec.~\ref{sec:alignment}: for each gate tap $x_i^{(\ell)}$ we compute
\[
s_i^{(\ell)} = \bigl|x_i^{(\ell)} \odot \partial \mathcal{L}_{\text{cls}} 
/ \partial x_i^{(\ell)}\bigr|,
\]
aggregate and normalize it within the layer, and match it to the corresponding \gate{} output $g_i^{(\ell)}(m)$. The total loss, $\mathcal{L}_{\text{total}}
= \mathcal{L}_{\text{cls}}
+ \alpha \Bigl( \mathcal{L}_{\text{align}}(g, s)
+ \lambda_{\text{bin}} \sum\nolimits_i g_i (1-g_i) \Bigr).$
\gate{} parameters receive gradients exclusively from the alignment term; the backbone is primarily optimized by $\mathcal{L}_{\text{cls}}$. This conditional training schedule prevents task optimization from being dominated by early pruning decisions, and allows \gate{} to learn a smooth, modality-conditioned saliency field that can later be queried without backpropagation. 


\subsection{Applying \gate{} for Structured Modality-aware Pruning}
\label{sec:pruning}



\begin{figure}
    \centering
    \includegraphics[width=\linewidth]{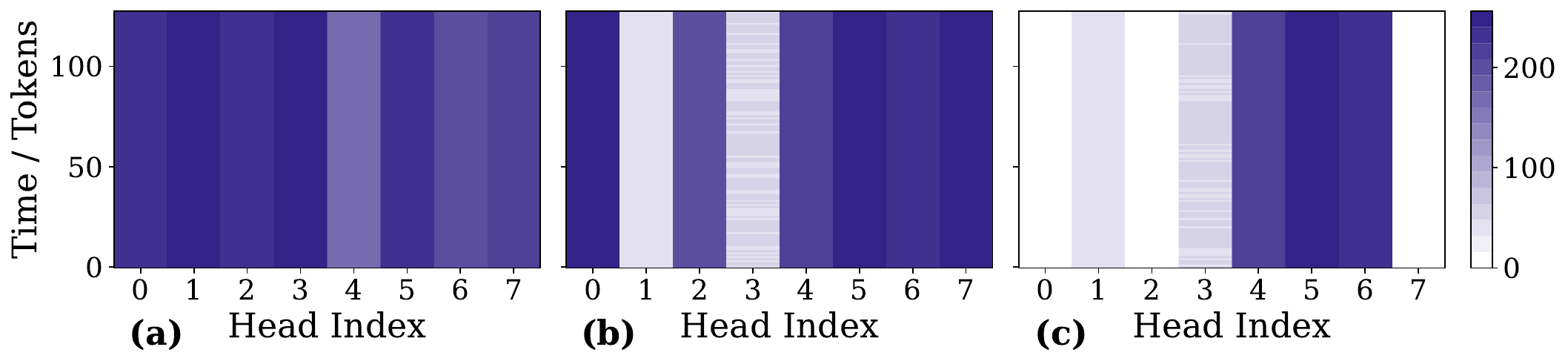}
    \vspace{-20pt}
    \caption{Attention weight heatmaps (head index vs.\ time/token) for a 
    a modality-encoder (wrist BVP from WESAD dataset~\cite{schmidt2018introducing}) at three stages. \textbf{(a)} Dense baseline (6.175~GFLOPs): attention is distributed uniformly across all 8 heads with low per-head magnitude. \textbf{(b)} Sparse attention prior to 
    \gate{} pruning (5.23~GFLOPs): inter-head variation emerges, with select heads becoming dominant, but all heads remain active. \textbf{(c)} \gate{}-pruned attention (3.81~GFLOPs, 28\% head pruning): redundant heads are fully gated out (white columns), concentrating computation on salient heads and timesteps while preserving discriminative temporal structure.}
    \label{fig:attn_triplet}
\end{figure}

Once \gate{} has been trained with the saliency alignment objective $\mathcal{L}_{\text{align}}$, each structural unit is now equipped with a modality-aware importance score. For a given deployment scenario, we fix the modality mask and evaluate the corresponding gate outputs $g_i^{(\ell)}(m_{\text{plat}})$ for all layers, then apply a global threshold determined by a desired pruning ratio to select a compact 
subnetwork. Unlike the traditional three-stage pruning pipeline (train--score--fine-tune), our approach performs task learning and structural selection under a unified training objective, so that once 
training converges, we can directly materialize a pruned model without additional fine-tuning.

We apply \gate{}-driven structured pruning at three locations: (1) multi-head self-attention heads, (2) feed-forward (FFN) layers in the Transformer encoders, and (3) internal FFN layers inside Sparse-MoE 
experts.

\smallskip
\noindent \textbf{Head pruning with GQA consistency.} In each attention layer, let $H_q$ denote the number of query heads and $H_k$ the number of key/value groups in the \attn{} topology. For a fixed platform mask, we aggregate \gate{} outputs over time and batch to obtain a mean gating score $\bar{g}_h$ for each query head $h$. We then apply a global threshold to obtain the retained head set $H_{\text{keep}} \subseteq \{0,\dots,H_q-1\}$. Given the query projection $W_Q \in \mathbb{R}^{d_{\text{model}} \times H_q d_h}$, we keep only the column blocks associated with $H_{\text{keep}}$. To preserve the \attn{} 
structure, we derive the set of key/value groups to keep via the head-to-group mapping, $
K_{\text{keep}} = \Big\lfloor \frac{H_{\text{keep}}}{H_q / H_k} \Big\rfloor,
$ and apply analogous block filtering to $W_K$ and $W_V$. If a key/value group is not referenced by any retained head, the entire group is removed.

\smallskip
\noindent\textbf{Feed-forward (FFN) pruning.}
Each Transformer encoder layer contains a feed-forward sublayer with two 
linear mappings $W_1 \in \mathbb{R}^{d_{\text{model}} \times d_{\text{ff}}}$ 
and $W_2 \in \mathbb{R}^{d_{\text{ff}} \times d_{\text{model}}}$. \gate{} 
provides an importance score for each hidden unit; we threshold these scores 
to obtain a retained unit set $N_{\text{keep}}$, and symmetrically 
reconstruct both weight matrices as, $W_1' = W_1[:, N_{\text{keep}}], \quad
W_2' = W_2[N_{\text{keep}}, :]$. This reduces the hidden dimension from $d_{\text{ff}}$ to $d_{\text{ff}}' 
= |N_{\text{keep}}|$, conditioned on the active modalities.

\smallskip
\noindent \textbf{Sparse-MoE FFN pruning.}
In the Sparse-MoE blocks, the total number of experts and Top-$k$ routing 
policy remain fixed. Each expert contains its own internal FFN; we reuse 
the \gate{} scores associated with expert-internal channels to prune these 
FFNs in the same symmetric manner, $
W_{1,e}' = W_{1,e}[:, N_{\text{keep},e}], \quad
W_{2,e}' = W_{2,e}[N_{\text{keep},e}, :],$ where $N_{\text{keep},e}$ is the retained unit set for expert $e$. Each 
expert independently shrinks its hidden width according to its learned 
saliency profile, while the global expert routing topology and Top-$k$ 
sparsity pattern remain intact.

Figure~\ref{fig:attn_triplet} illustrates the effect of \gate{}-driven head 
pruning on a representative wrist-BVP attention layer. The left panel 
(\emph{Dense}) shows the multi-head attention heatmap at $6.18$ GFLOPs. The 
middle panel (\emph{Sparse}) corresponds to \attn{} before pruning, already 
reducing the layer cost to $5.23$ GFLOPs. The right panel (\emph{Pruned}) 
shows the same layer after \gate{}-based head pruning at $28\%$, further 
reducing the cost to $3.81$ GFLOPs, while preserving salient temporal 
patterns.

\section{Experimental Setup} \label{sec:exp}

\noindent\textbf{Backbones and datasets.} We evaluate \method{} on three multimodal time-series benchmarks: WESAD \cite{schmidt2018introducing}, DaliaHAR \cite{reiss2019deep}, and DSADS \cite{altun2010comparative}, using three representative multimodal backbones: FlexMoE \cite{yun2024flex}, FuseMoE \cite{NEURIPS2024_7d62a85e}, and MAESTRO \cite{mohapatra2025maestro}. All backbones are trained and deployed with their official default settings, and \gate{} is trained on an NVIDIA L40.


\smallskip
\noindent\textbf{Baselines.} For structured pruning, we compare \gate{} against three baselines: Random (uniform random removal), Magnitude (pruning smallest-magnitude units), and SynFlow~\cite{tanaka2020synflow} (data-free synaptic saliency scoring). For efficient attention, we compare each backbone against its \attn{} variant, in which dense self-attention is replaced by \attn{}. For end-to-end evaluation, we assess three configurations: the original backbone, the backbone with \gate{}, and the full \method{} framework combining both \gate{} and \attn{}.


\smallskip
\noindent\textbf{Comparison settings.} We conduct several sets of experiments. First, we compare the four pruning methods mentioned above under full-modality and modality-dropout settings across all backbone-dataset combinations. Next, we evaluate \attn{} on two standard time-series backbones, Transformer~\cite{vaswani2017attention} and iTransformer~\cite{liu2023itransformer}, as well as the three multimodal backbones, assessing predictive performance and computational efficiency. We then evaluate the integrated \method{} design on the best backbone through cross-platform hardware deployment and post-training quantization. Finally, we perform ablations on the saliency teacher and grouped-query attention configuration. 

\smallskip
\noindent\textbf{Evaluation metrics.} We report accuracy on the evaluation split. We report GFLOPs as the forward-pass floating-point operation count per inference. FLOP statistics are measured primarily with \texttt{fvcore}~\cite{fvcore} and, where needed, THOP~\cite{thop}. For deployment experiments, we report memory as serialized checkpoint size and latency as end-to-end wall-clock inference time. For post-training quantization, we additionally compare accuracy and model storage across numeric precisions.


\smallskip
\noindent\textbf{Platforms.} We benchmark all configurations on heterogeneous hardware, including an NVIDIA L40 GPU, a CPU platform, a Jetson TX2, a Google Pixel 8, and an iPhone 13 Pro Max. Server experiments run in PyTorch~2.5.1 with CUDA~12.1~\cite{pytorch}, whereas Jetson TX2 experiments use PyTorch~1.8.0. Mobile deployment uses ExecuTorch~1.1~\cite{executorch}.


\section{Evaluation}
\label{sec:eval}

\subsection{Modality-aware Pruning under Missingness}

\begin{figure}[!htbp]
    \centering
    \includegraphics[width=\linewidth]{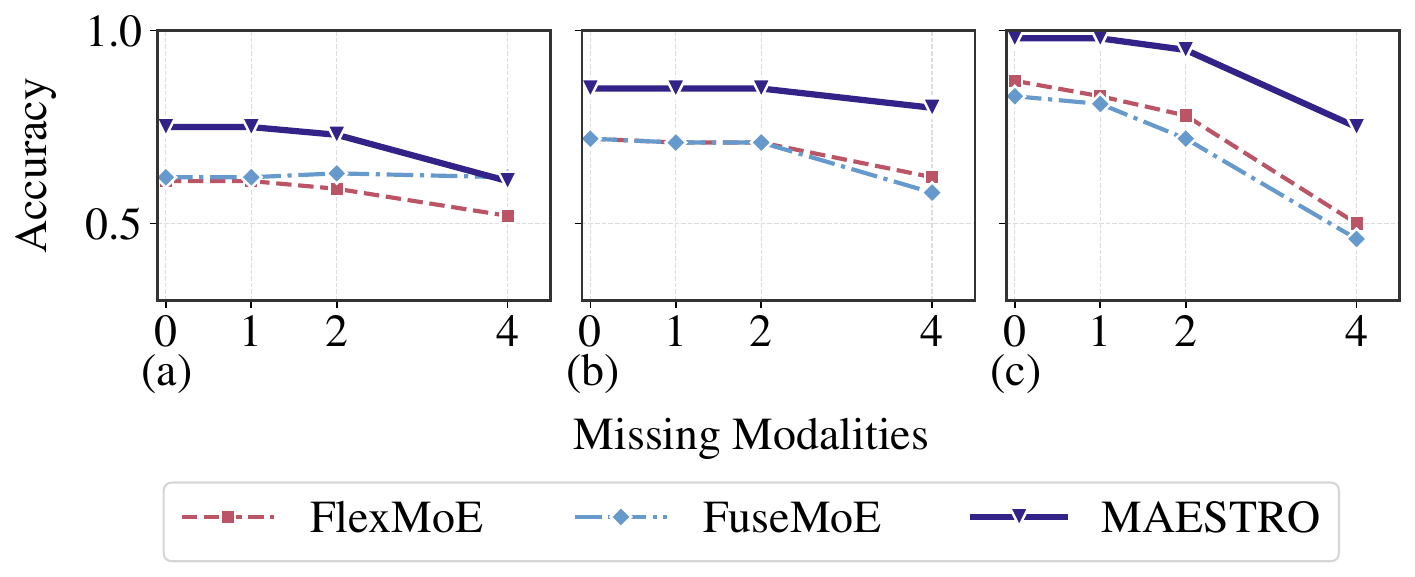}
    \vspace{-20pt}
    \caption{ Accuracy of FlexMoE+\gate{}, FuseMoE+\gate{}, and MAESTRO+\gate{} under different numbers of missing modalities (0, 1, 2, and 4), on (a) WESAD, (b) DaliaHAR, and (c) DSADS.}
    \vspace{-10pt}
    \label{fig:missingness}
\end{figure}

We evaluate modality-aware pruning under missingness both before and after structured pruning. Figure~\ref{fig:missingness} reports the accuracy of the unpruned backbones equipped with \gate{}, and Tables~\ref{tab:multimodal_backbone_pruning_acc_wesad}--\ref{tab:multimodal_backbone_pruning_acc_dsads} report accuracy after structured pruning using \gate{}-predicted importance scores across pruning ratios and modality-dropout settings.

\begin{table*}[!htbp]
\centering
\setlength{\tabcolsep}{2.1pt}
\renewcommand{\arraystretch}{1.00}
\caption{WESAD accuracy under modality dropout and pruning ratios of 6\%, 12\%, 17\%, and 23\%. \localbest{Orange} marks backbone-wise local best; \best{bold orange} marks global best.}
\label{tab:multimodal_backbone_pruning_acc_wesad}
\vspace{-4pt}
\begin{tabular}{@{}llcccccccccccccccc@{}}
\toprule
\multirow{2}{*}{Backbone}
& \multirow{2}{*}{Strategy}
& \multicolumn{4}{c}{Full}
& \multicolumn{4}{c}{Drop 1}
& \multicolumn{4}{c}{Drop 2}
& \multicolumn{4}{c}{Drop 4} \\
\cmidrule(lr){3-6}\cmidrule(lr){7-10}\cmidrule(lr){11-14}\cmidrule(lr){15-18}
& & 6 & 12 & 17 & 23 & 6 & 12 & 17 & 23 & 6 & 12 & 17 & 23 & 6 & 12 & 17 & 23 \\
\midrule
\multirow{4}{*}{FlexMoE}
& Random
  & \localbest{0.62} & \localbest{0.62} & 0.59 & 0.63
  & \localbest{0.62} & \localbest{0.62} & 0.58 & 0.59
  & \localbest{0.61} & \localbest{0.61} & 0.56 & 0.55
  & 0.53 & 0.49 & 0.43 & 0.37 \\
& Magnitude
  & 0.61 & 0.61 & 0.56 & 0.55
  & 0.61 & 0.61 & 0.56 & 0.55
  & 0.59 & 0.59 & 0.56 & 0.55
  & 0.51 & 0.51 & 0.46 & 0.46 \\
& Synflow
  & 0.56 & 0.56 & 0.54 & 0.54
  & 0.56 & 0.56 & 0.54 & 0.55
  & 0.53 & 0.54 & 0.38 & 0.38
  & \localbest{0.55} & \localbest{0.52} & 0.49 & 0.49 \\
& \gate{}
  & 0.61 & 0.61 & \localbest{0.62} & \localbest{0.65}
  & 0.61 & 0.61 & \localbest{0.62} & \localbest{0.65}
  & 0.59 & 0.59 & \localbest{0.60} & \localbest{0.63}
  & 0.52 & \localbest{0.52} & \localbest{0.52} & \localbest{0.54} \\
\midrule
\multirow{4}{*}{FuseMoE}
& Random
  & 0.59 & 0.59 & 0.57 & 0.57
  & 0.59 & 0.59 & 0.57 & 0.57
  & 0.51 & 0.48 & 0.49 & 0.46
  & 0.61 & 0.58 & 0.59 & 0.53 \\
& Magnitude
  & 0.59 & 0.57 & 0.57 & 0.57
  & 0.59 & 0.57 & 0.57 & 0.57
  & 0.57 & 0.58 & 0.57 & 0.57
  & 0.53 & 0.53 & 0.53 & 0.53 \\
& Synflow
  & 0.60 & 0.59 & 0.59 & 0.59
  & 0.59 & 0.58 & 0.58 & 0.58
  & 0.59 & 0.57 & 0.57 & 0.57
  & 0.61 & 0.60 & 0.60 & 0.60 \\
& \gate{}
  & \localbest{0.61} & \localbest{0.61} & \localbest{0.62} & \localbest{0.65}
  & \localbest{0.62} & \localbest{0.62} & \localbest{0.61} & \localbest{0.61}
  & \localbest{0.62} & \localbest{0.62} & \localbest{0.62} & \localbest{0.62}
  & \best{0.62} & \best{0.62} & \best{0.62} & \best{0.62} \\
\midrule
\multirow{4}{*}{MAESTRO}
& Random
  & 0.74 & 0.73 & 0.63 & 0.52
  & 0.74 & 0.73 & 0.64 & 0.54
  & \best{0.73} & 0.71 & 0.63 & 0.55
  & \localbest{0.61} & 0.60 & 0.55 & 0.51 \\
& Magnitude
  & \best{0.75} & 0.70 & 0.64 & 0.59
  & 0.74 & 0.70 & 0.64 & 0.59
  & 0.72 & 0.68 & 0.62 & 0.59
  & \localbest{0.61} & 0.59 & 0.58 & 0.56 \\
& Synflow
  & \best{0.75} & \best{0.75} & 0.71 & 0.72
  & \best{0.75} & \best{0.75} & 0.71 & 0.71
  & \best{0.73} & \best{0.72} & 0.72 & 0.71
  & \localbest{0.61} & \localbest{0.61} & \best{0.62} & \best{0.62} \\
& \gate{}
  & \best{0.75} & \best{0.75} & \best{0.74} & \best{0.75}
  & \best{0.75} & \best{0.75} & \best{0.74} & \best{0.75}
  & \best{0.73} & \best{0.72} & \best{0.73} & \best{0.73}
  & \localbest{0.61} & \localbest{0.61} & 0.61 & 0.61 \\
\bottomrule
\end{tabular}
\end{table*}

\begin{table*}[!htbp]
\centering
\setlength{\tabcolsep}{2.1pt}
\renewcommand{\arraystretch}{1.00}
\caption{DaliaHAR accuracy under modality dropout and pruning ratios of 6\%, 12\%, 17\%, and 23\%. \localbest{Orange} marks backbone-wise local best; \best{bold orange} marks global best.}
\label{tab:multimodal_backbone_pruning_acc_dalia}
\vspace{-4pt}
\begin{tabular}{@{}llcccccccccccccccc@{}}
\toprule
\multirow{2}{*}{Backbone}
& \multirow{2}{*}{Strategy}
& \multicolumn{4}{c}{Full}
& \multicolumn{4}{c}{Drop 1}
& \multicolumn{4}{c}{Drop 2}
& \multicolumn{4}{c}{Drop 4} \\
\cmidrule(lr){3-6}\cmidrule(lr){7-10}\cmidrule(lr){11-14}\cmidrule(lr){15-18}
& & 6 & 12 & 17 & 23 & 6 & 12 & 17 & 23 & 6 & 12 & 17 & 23 & 6 & 12 & 17 & 23 \\
\midrule
\multirow{4}{*}{FlexMoE}
& Random
  & 0.58 & 0.47 & 0.43 & 0.22
  & 0.57 & 0.49 & 0.47 & \localbest{0.23}
  & 0.57 & 0.48 & 0.46 & 0.22
  & 0.39 & 0.30 & 0.30 & \localbest{0.40} \\
& Magnitude
  & 0.47 & 0.36 & 0.28 & 0.18
  & 0.48 & 0.38 & 0.31 & 0.19
  & 0.47 & 0.36 & 0.30 & 0.19
  & 0.41 & 0.38 & 0.28 & 0.17 \\
& Synflow
  & 0.61 & 0.57 & 0.35 & 0.17
  & 0.62 & 0.58 & 0.37 & 0.17
  & 0.61 & 0.58 & 0.38 & 0.16
  & 0.58 & 0.53 & 0.32 & 0.16 \\
& \gate{}
  & \localbest{0.72} & \localbest{0.68} & \localbest{0.57} & \localbest{0.24}
  & \localbest{0.71} & \localbest{0.68} & \localbest{0.56} & \localbest{0.23}
  & \localbest{0.71} & \localbest{0.66} & \localbest{0.56} & \localbest{0.34}
  & \localbest{0.61} & \localbest{0.55} & \localbest{0.44} & 0.36 \\
\midrule
\multirow{4}{*}{FuseMoE}
& Random
  & 0.49 & 0.60 & 0.47 & 0.44
  & 0.50 & 0.61 & 0.46 & 0.43
  & 0.47 & 0.56 & 0.42 & 0.39
  & 0.40 & 0.22 & 0.34 & 0.31 \\
& Magnitude
  & 0.66 & 0.54 & 0.50 & 0.47
  & 0.67 & 0.54 & 0.50 & 0.67
  & 0.66 & 0.52 & 0.49 & 0.45
  & \localbest{0.58} & 0.51 & 0.35 & 0.33 \\
& Synflow
  & 0.65 & 0.57 & 0.55 & 0.52
  & 0.65 & 0.51 & 0.50 & 0.48
  & 0.65 & 0.55 & 0.54 & 0.53
  & 0.49 & 0.49 & 0.50 & 0.51 \\
& \gate{}
  & \localbest{0.72} & \localbest{0.72} & \localbest{0.72} & \localbest{0.71}
  & \localbest{0.71} & \localbest{0.71} & \localbest{0.71} & \localbest{0.68}
  & \localbest{0.71} & \localbest{0.71} & \localbest{0.70} & \localbest{0.67}
  & \localbest{0.58} & \localbest{0.59} & \localbest{0.60} & \localbest{0.58} \\
\midrule
\multirow{4}{*}{MAESTRO}
& Random
  & 0.83 & 0.83 & 0.80 & 0.64
  & 0.84 & 0.83 & 0.79 & 0.63
  & 0.83 & 0.83 & 0.79 & 0.64
  & 0.77 & 0.75 & 0.66 & 0.48 \\
& Magnitude
  & 0.80 & 0.79 & 0.74 & 0.73
  & 0.80 & 0.78 & 0.75 & 0.74
  & 0.80 & 0.78 & 0.74 & 0.73
  & 0.73 & 0.68 & 0.55 & 0.53 \\
& Synflow
  & \best{0.85} & 0.78 & 0.75 & 0.74
  & \best{0.85} & 0.79 & 0.77 & 0.73
  & \best{0.85} & 0.80 & 0.78 & 0.74
  & \best{0.79} & 0.73 & 0.67 & 0.61 \\
& \gate{}
  & \best{0.85} & \best{0.85} & \best{0.85} & \best{0.85}
  & \best{0.85} & \best{0.85} & \best{0.85} & \best{0.85}
  & \best{0.85} & \best{0.85} & \best{0.85} & \best{0.85}
  & \best{0.79} & \best{0.79} & \best{0.80} & \best{0.79} \\
\bottomrule
\end{tabular}
\end{table*}

\begin{table*}[!htbp]
\centering
\setlength{\tabcolsep}{2.1pt}
\renewcommand{\arraystretch}{1.00}
\caption{DSADS accuracy under modality dropout and pruning ratios of 6\%, 12\%, 17\%, and 23\%. \localbest{Orange} marks backbone-wise local best; \best{bold orange} marks global best.}
\label{tab:multimodal_backbone_pruning_acc_dsads}
\vspace{-4pt}
\begin{tabular}{@{}llcccccccccccccccc@{}}
\toprule
\multirow{2}{*}{Backbone}
& \multirow{2}{*}{Strategy}
& \multicolumn{4}{c}{Full}
& \multicolumn{4}{c}{Drop 1}
& \multicolumn{4}{c}{Drop 2}
& \multicolumn{4}{c}{Drop 4} \\
\cmidrule(lr){3-6}\cmidrule(lr){7-10}\cmidrule(lr){11-14}\cmidrule(lr){15-18}
& & 6 & 12 & 17 & 23 & 6 & 12 & 17 & 23 & 6 & 12 & 17 & 23 & 6 & 12 & 17 & 23 \\
\midrule
\multirow{4}{*}{FlexMoE}
& Random
  & 0.81 & 0.13 & 0.09 & 0.07
  & 0.78 & 0.12 & 0.11 & 0.09
  & 0.64 & 0.14 & 0.11 & 0.10
  & 0.41 & 0.08 & 0.08 & 0.08 \\
& Magnitude
  & 0.84 & 0.55 & 0.16 & 0.11
  & 0.82 & 0.52 & 0.19 & 0.14
  & 0.73 & 0.34 & 0.14 & 0.12
  & 0.46 & 0.22 & 0.16 & 0.10 \\
& Synflow
  & 0.84 & 0.52 & 0.18 & 0.13
  & 0.80 & 0.60 & 0.14 & 0.12
  & 0.71 & 0.44 & 0.16 & \localbest{0.14}
  & 0.40 & 0.30 & 0.11 & 0.09 \\
& \gate{}
  & \localbest{0.86} & \localbest{0.86} & \localbest{0.57} & \localbest{0.23}
  & \localbest{0.83} & \localbest{0.81} & \localbest{0.49} & \localbest{0.19}
  & \localbest{0.78} & \localbest{0.75} & \localbest{0.42} & 0.12
  & \localbest{0.49} & \localbest{0.45} & \localbest{0.21} & \localbest{0.16} \\
\midrule
\multirow{4}{*}{FuseMoE}
& Random
  & 0.81 & 0.49 & 0.77 & 0.74
  & 0.75 & 0.73 & 0.71 & 0.70
  & 0.70 & 0.66 & 0.63 & 0.55
  & 0.46 & 0.39 & 0.36 & 0.31 \\
& Magnitude
  & \localbest{0.83} & 0.81 & 0.81 & 0.81
  & \localbest{0.81} & \localbest{0.81} & 0.80 & 0.80
  & 0.68 & 0.66 & 0.66 & 0.66
  & 0.45 & 0.41 & 0.40 & 0.41 \\
& Synflow
  & \localbest{0.83} & 0.81 & 0.81 & 0.80
  & 0.80 & 0.78 & 0.78 & 0.77
  & 0.70 & 0.62 & 0.62 & 0.62
  & 0.36 & 0.33 & 0.31 & 0.31 \\
& \gate{}
  & \localbest{0.83} & \localbest{0.83} & \localbest{0.83} & \localbest{0.83}
  & \localbest{0.81} & \localbest{0.81} & \localbest{0.81} & \localbest{0.81}
  & \localbest{0.72} & \localbest{0.72} & \localbest{0.71} & \localbest{0.71}
  & \localbest{0.47} & \localbest{0.46} & \localbest{0.46} & \localbest{0.45} \\
\midrule
\multirow{4}{*}{MAESTRO}
& Random
  & \best{0.98} & \best{0.98} & \best{0.98} & \best{0.97}
  & 0.97 & 0.96 & 0.94 & 0.88
  & 0.92 & 0.90 & 0.85 & 0.82
  & 0.71 & 0.68 & 0.63 & 0.60 \\
& Magnitude
  & 0.97 & \best{0.98} & 0.97 & \best{0.97}
  & 0.97 & 0.94 & 0.92 & 0.90
  & \best{0.95} & 0.92 & 0.89 & 0.89
  & 0.74 & 0.72 & 0.68 & 0.65 \\
& Synflow
  & \best{0.98} & 0.97 & 0.96 & 0.95
  & 0.97 & 0.97 & 0.96 & 0.92
  & \best{0.95} & \best{0.94} & 0.91 & 0.87
  & \best{0.75} & 0.67 & 0.62 & 0.60 \\
& \gate{}
  & \best{0.98} & \best{0.98} & \best{0.98} & \best{0.97}
  & \best{0.98} & \best{0.98} & \best{0.97} & \best{0.97}
  & \best{0.95} & \best{0.94} & \best{0.94} & \best{0.91}
  & \best{0.75} & \best{0.73} & \best{0.73} & \best{0.72} \\
\bottomrule
\end{tabular}
\end{table*}

Across the 144 backbone--dataset--pruning--missingness combinations in Tables~\ref{tab:multimodal_backbone_pruning_acc_wesad}--\ref{tab:multimodal_backbone_pruning_acc_dsads}, \gate{} matches or outperforms the strongest modality-agnostic baseline in 133 cases (92.4\%). Averaged over all backbones, datasets, pruning ratios, and missingness settings, \gate{} achieves 0.68 accuracy, compared with 0.60 for the strongest competing pruning baseline on average (SynFlow), corresponding to a 12.7\% relative improvement. The gain appears on both strong and weak backbones. For example, on MAESTRO over DaliaHAR at 23\% pruning with four missing modalities, \gate{} improves accuracy from 0.61 to 0.79; on FlexMoE over DSADS at 17\% pruning with full modalities, it improves accuracy from 0.18 to 0.57, yielding a 3.2$\times$ gain over the strongest baseline. These results show that the modality-conditioned importance scores generated by \gate{} preserve accuracy more effectively than static pruning rules.

The improvement becomes even larger under severe missingness. With four modalities dropped, the average accuracy of \gate{} across all backbones, datasets, and pruning ratios is 0.57, compared with 0.50 for the second-best pruning baseline on average, yielding a 13.4\% relative improvement. The same trend is visible in the hardest per-dataset settings: at 23\% pruning with four missing modalities, \gate{} improves accuracy from 0.61 to 0.79 on DaliaHAR and from 0.65 to 0.72 on DSADS. As sensing conditions become more adverse, the benefit of modality-aware pruning becomes increasingly pronounced.

Figure~\ref{fig:missingness} further shows that the benefit of \gate{} can be amplified when it is paired with a stronger multimodal backbone. Across missingness levels from 0 to 4, MAESTRO+\gate{} maintains accuracy of 0.85/0.85/0.85/0.80 on DaliaHAR and 0.98/0.98/0.95/0.75 on DSADS before pruning, while FlexMoE+\gate{} and FuseMoE+\gate{} degrade more noticeably. Together, these results indicate that \gate{} consistently improves robustness under missingness, and that stronger multimodal backbones can further extend this gain.

\subsection{Efficiency Gains Using \attn{}}

Table~\ref{tab:attn_efficiency_all} evaluates the effect of replacing dense self-attention with \attn{} across different backbones and datasets. We compare the original models and their \attn{} variants in terms of predictive performance and computational cost measured by GFLOPs.

\begin{table}[!htbp]
\centering
\setlength{\tabcolsep}{3.0pt}
\renewcommand{\arraystretch}{0.88}
\caption{Accuracy and GFLOPs before and after \attn{} on WESAD, DaliaHAR, and DSADS. Higher accuracy and lower GFLOPs are better; best and second-best values are shown in \textcolor{LinkColor}{bold orange} and underlined, respectively.}
\label{tab:attn_efficiency_all}
\vspace{-3pt}
\begin{tabular}{@{}llcccc@{}}
\toprule
\multirow{2}{*}{Dataset} & \multirow{2}{*}{Backbone}
& \multicolumn{2}{c}{Orig.}
& \multicolumn{2}{c}{+\attn{}} \\
\cmidrule(lr){3-4}\cmidrule(lr){5-6}
& & Acc$\uparrow$ & GFLOPs$\downarrow$ & Acc$\uparrow$ & GFLOPs$\downarrow$ \\
\midrule
\multirow{5}{*}{WESAD}
& MAESTRO      & \best{0.75} & \secondbest{6.15} & \best{0.76} & \secondbest{5.23} \\
& Transformer  & 0.67 & 22.25 & 0.65 & 15.81 \\
& iTransformer & 0.61 & \best{5.76} & 0.62 & \best{5.06} \\
& FlexMoE      & 0.65 & 16.68 & \secondbest{0.67} & 16.35 \\
& FuseMoE      & \secondbest{0.68} & 12.28 & 0.65 & 11.47 \\
\midrule
\multirow{5}{*}{DaliaHAR}
& MAESTRO      & \best{0.85} & \best{3.07} & \best{0.82} & \best{2.60} \\
& Transformer  & \secondbest{0.71} & 22.23 & 0.71 & 15.79 \\
& iTransformer & 0.63 & \secondbest{3.70} & 0.64 & \secondbest{3.25} \\
& FlexMoE      & 0.70 & 8.35 & 0.71 & 8.17 \\
& FuseMoE      & 0.70 & 12.26 & \secondbest{0.73} & 11.44 \\
\midrule
\multirow{5}{*}{DSADS}
& MAESTRO      & \best{0.98} & \best{1.53} & \best{0.88} & \best{1.30} \\
& Transformer  & 0.78 & 11.02 & \secondbest{0.83} & 7.84 \\
& iTransformer & 0.51 & 18.44 & 0.51 & 16.18 \\
& FlexMoE      & \secondbest{0.84} & 12.11 & 0.80 & 11.44 \\
& FuseMoE      & 0.83 & \secondbest{5.99} & 0.82 & \secondbest{5.59} \\
\bottomrule
\end{tabular}
\end{table}

A first clear trend is that \attn{} consistently reduces model GFLOPs, confirming its effectiveness as an efficient attention replacement. Across all evaluated backbones, applying \attn{} lowers GFLOPs on WESAD, DaliaHAR, and DSADS, with the largest reductions observed on Transformer, where GFLOPs decrease from 22.25 to 15.81 on WESAD, from 22.23 to 15.79 on DaliaHAR, and from 11.02 to 7.84 on DSADS, corresponding to about 29.0\% savings. Similar trends also hold for multimodal backbones. For example, MAESTRO reduces GFLOPs from 6.15 to 5.23 on WESAD, from 3.07 to 2.60 on DaliaHAR, and from 1.53 to 1.30 on DSADS, yielding a stable $\sim$15\% reduction. Moreover, \attn{} can improve efficiency without sacrificing predictive performance, and in some cases even improves both. On WESAD, FlexMoE improves accuracy from 0.65 to 0.67 while reducing GFLOPs from 16.68 to 16.35. On DaliaHAR, FuseMoE improves accuracy from 0.70 to 0.73 while lowering GFLOPs from 12.26 to 11.44. On DSADS, Transformer improves accuracy from 0.78 to 0.83 while reducing GFLOPs from 11.02 to 7.84. These results show that \attn{} improves attention efficiency across both vanilla Transformer-style models and multimodal backbones.

MAESTRO remains the strongest backbone after introducing \attn{}, delivering the best overall accuracy while maintaining the lowest or near-lowest GFLOPs among the compared models. With \attn{}, MAESTRO achieves 0.76 accuracy at 5.23 GFLOPs on WESAD, 0.82 accuracy at 2.60 GFLOPs on DaliaHAR, and 0.88 accuracy at 1.30 GFLOPs on DSADS. Compared with the best baseline equipped with \attn{}, MAESTRO improves accuracy by 13.4\% on WESAD (0.76 vs.~0.67), 12.3\% on DaliaHAR (0.82 vs.~0.73), and 6.0\% on DSADS (0.88 vs.~0.83), while requiring 3.1$\times$, 4.4$\times$, and 6.0$\times$ fewer GFLOPs, respectively. Although MAESTRO incurs some accuracy loss on DaliaHAR and DSADS relative to its original version (0.85 to 0.82 and 0.98 to 0.88), it still remains the highest-accuracy \attn{} backbone by a clear margin while being substantially more efficient than the alternatives.

Overall, these results support two conclusions. First, \attn{} consistently improves computational efficiency by reducing attention-related GFLOPs across diverse backbones, and can even improve accuracy in several cases. Second, MAESTRO remains the strongest backbone for this efficient-attention mechanism, which motivates our use of MAESTRO as the backbone in the subsequent integrated evaluation.

\subsection{Performance Gains Across Various Hardware Using \method{}}

Table~\ref{tab:hardware_perf_wesad} compares the original MAESTRO model, MAESTRO equipped with \gate{}, and then MAESTRO with the full \method{} framework, i.e., MAESTRO + \gate{} + \attn{} with 23\% pruning, across heterogeneous hardware platforms. There are two distinct trends. \textbf{First, the complete \method{} configuration consistently improves predictive performance while further reducing deployment cost.} Relative to the original MAESTRO backbone, \textbf{\method{} reduces model memory from 6.07 MB to 4.36 MB and GFLOPs from 6.83 to 4.41, corresponding to 28.2\% and 35.4\% reductions, respectively, while increasing accuracy from 0.75 to 0.77 on GPU and to 0.76 on CPU, Jetson, iPhone 13 Pro, and Google Pixel 8.} Even compared with the already pruned MAESTRO+\gate{} variant, \method{} further reduces memory by 9.2\% and GFLOPs by 13.0\%, while improving accuracy by 0.02 on every evaluated platform. This result shows that integrating \attn{} on top of \gate{} yields a strictly better accuracy-efficiency trade-off than pruning alone.

\begin{table}[H]
\centering
\setlength{\tabcolsep}{2.6pt}
\renewcommand{\arraystretch}{0.84}
\caption{WESAD hardware deployment (MAESTRO; latter two use 23\% pruning; \method{}=\gate{}+\attn{}).}
\label{tab:hardware_perf_wesad}
\vspace{-3pt}
\begin{tabular}{@{}lccc@{}}
\toprule
Metric & MAESTRO & +\gate{} & +\method{} \\
\midrule
Mem. (MB)$\downarrow$ & 6.07 & 4.80 & 4.36 \\
GFLOPs$\downarrow$    & 6.83 & 5.07 & 4.41 \\
\midrule
\multicolumn{4}{c}{Accuracy / Latency (ms)$\downarrow$} \\
\midrule
GPU       & 0.75/24.50  & 0.75/24.35  & 0.77/24.32 \\
CPU       & 0.74/60.19  & 0.74/43.96  & 0.76/42.72 \\
Jetson    & 0.74/295.87 & 0.74/257.82 & 0.76/254.77 \\
iPhone 13 & 0.75/167.82 & 0.74/121.07 & 0.76/113.32 \\
Pixel 8   & 0.75/256.28 & 0.74/169.96 & 0.76/157.43 \\
\bottomrule
\end{tabular}
\end{table}

\begin{figure}[!htbp]
    \centering
    \includegraphics[width=\linewidth]{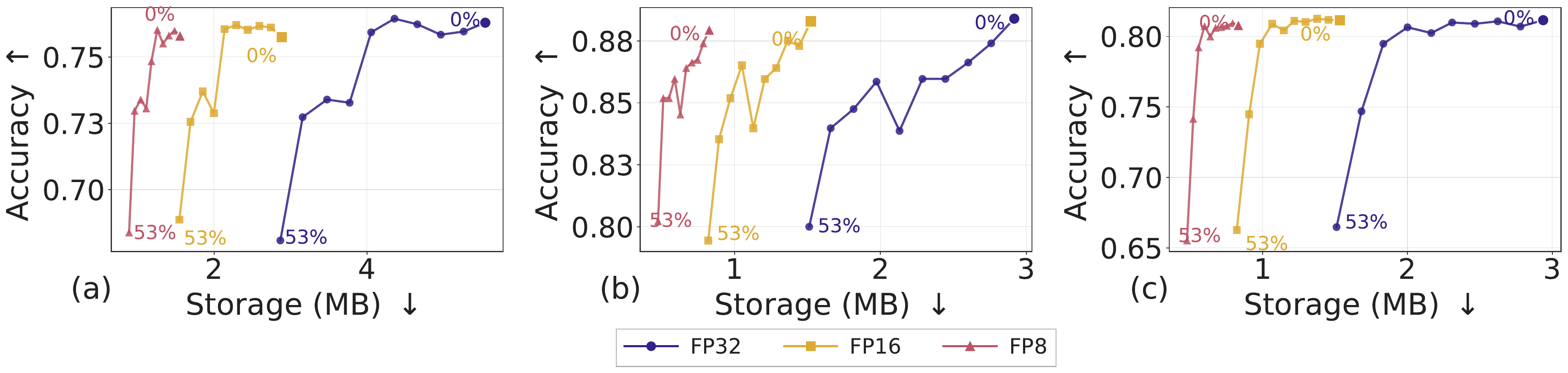}
    \vspace{-18pt}
    \caption{Accuracy–storage Pareto trade-offs for \method{} under post-training quantization on (a) WESAD, (b) DSADS and (c) DaliaHAR. Each point represents a model with a specific pruning ratio and numerical precision (FP32, FP16, FP8). The dashed line indicates the Pareto frontier across all configurations.}
    \label{fig:pareto}
\end{figure}

\begin{figure}[!htbp]
    \vspace{0pt}
    \centering
    \includegraphics[width=1.0\linewidth]{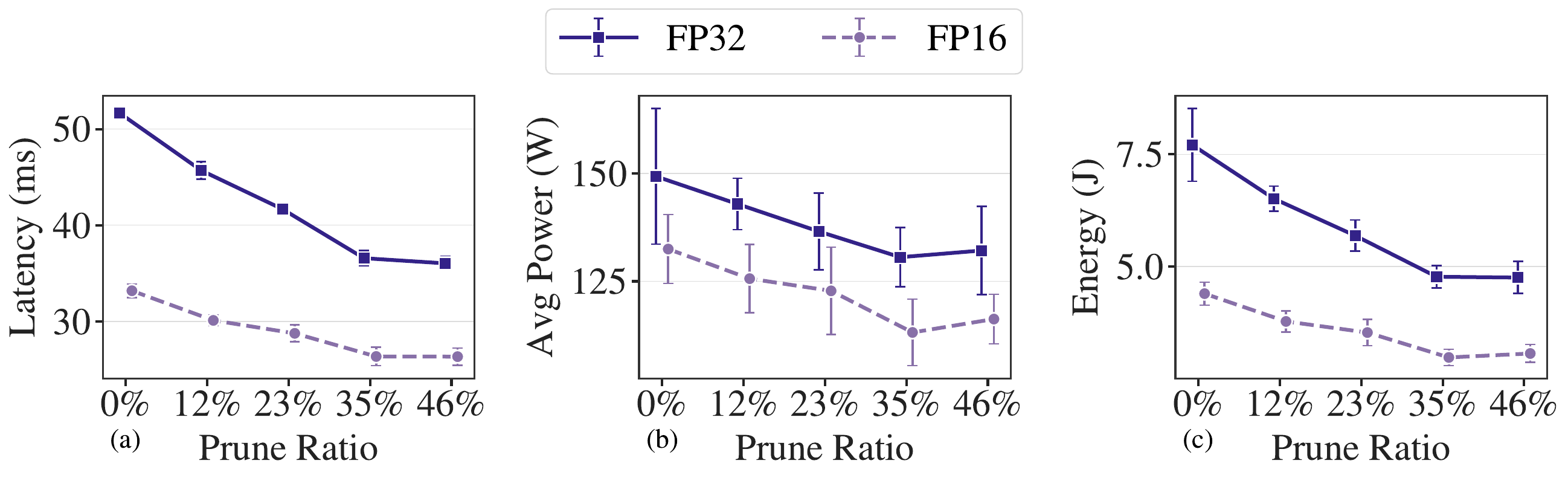}
    \vspace{-25pt}
    \caption{Latency, average power draw, and total energy consumption 
of \method{} under FP32 and FP16 quantization across pruning ratios.}

    \vspace{0pt}
    \label{fig:power_flow}
\end{figure}

\smallskip
\noindent\textbf{Compatibility with quantization.} We apply FP32, FP16, and FP8 post-training quantization on top of \method{}-pruned models, treating precision and pruning ratio as two independent compression axes. Figure~\ref{fig:pareto} shows the resulting Pareto frontier of accuracy versus storage across all three datasets. Three observations stand out. First, FP8 dominates the storage-constrained regime: across all datasets, FP8 at moderate pruning sits on or near the Pareto frontier, matching
unpruned FP32 accuracy at up to $5\times$ lower storage on WESAD. Second, quantization precision has negligible impact on accuracy below 28\% pruning. Third, the accuracy drop at high pruning ratios (e.g., 50\%) is consistent across all three precisions, indicating the bottleneck is structural rather than numerical. Figure~\ref{fig:power_flow} shows that FP16 quantization reduces latency, power, and energy by up to 35\% relative to FP32, with 
gains compounding at higher pruning ratios showing the compatibility of \method{} with other model compression techniques.

\subsection{Ablation Study}


\begin{figure}[!htbp]
    \vspace{0pt}
    \centering
    \includegraphics[width=1.0\linewidth]{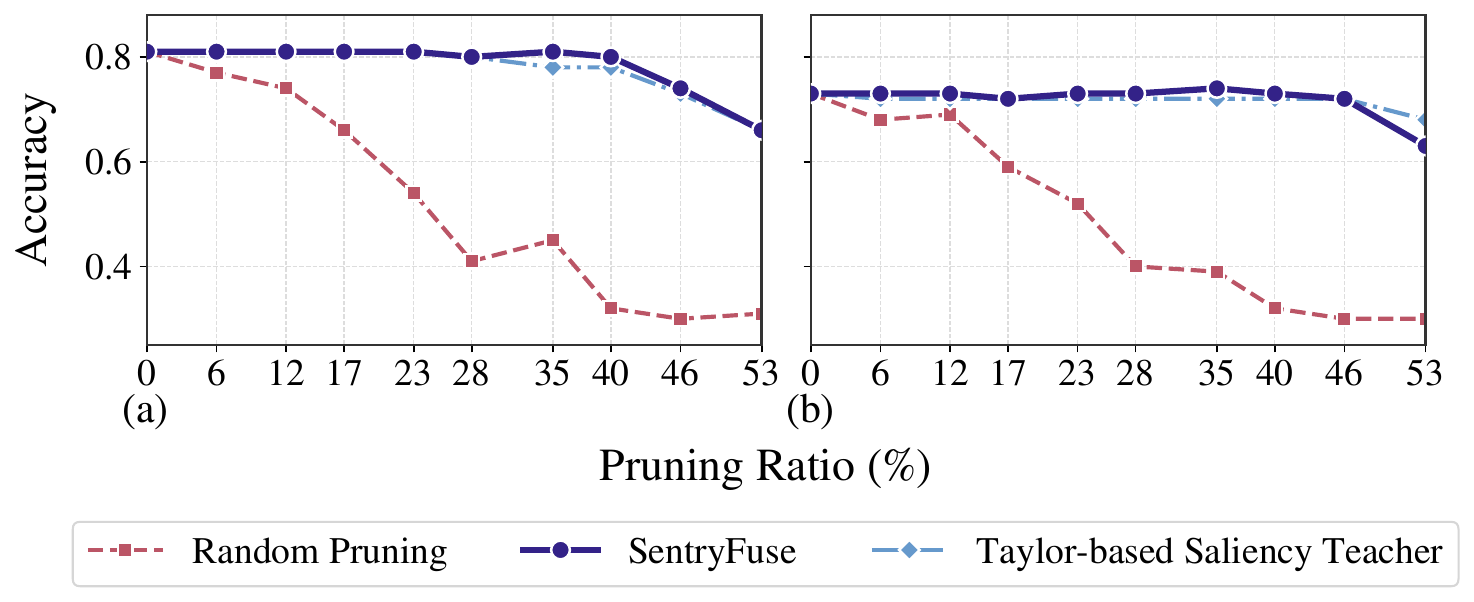}
    \vspace{-25pt}
    \caption{Accuracy of random pruning, our \method{}, and a Taylor-based saliency teacher across pruning ratios on DaliaHAR under (a) full modalities and (b) 4-modality dropout. }

    \vspace{0pt}
    \label{fig:bound}
\end{figure}

\begin{figure}[!htbp]
    \vspace{0pt}
    \centering
    \includegraphics[width=1.0\linewidth]{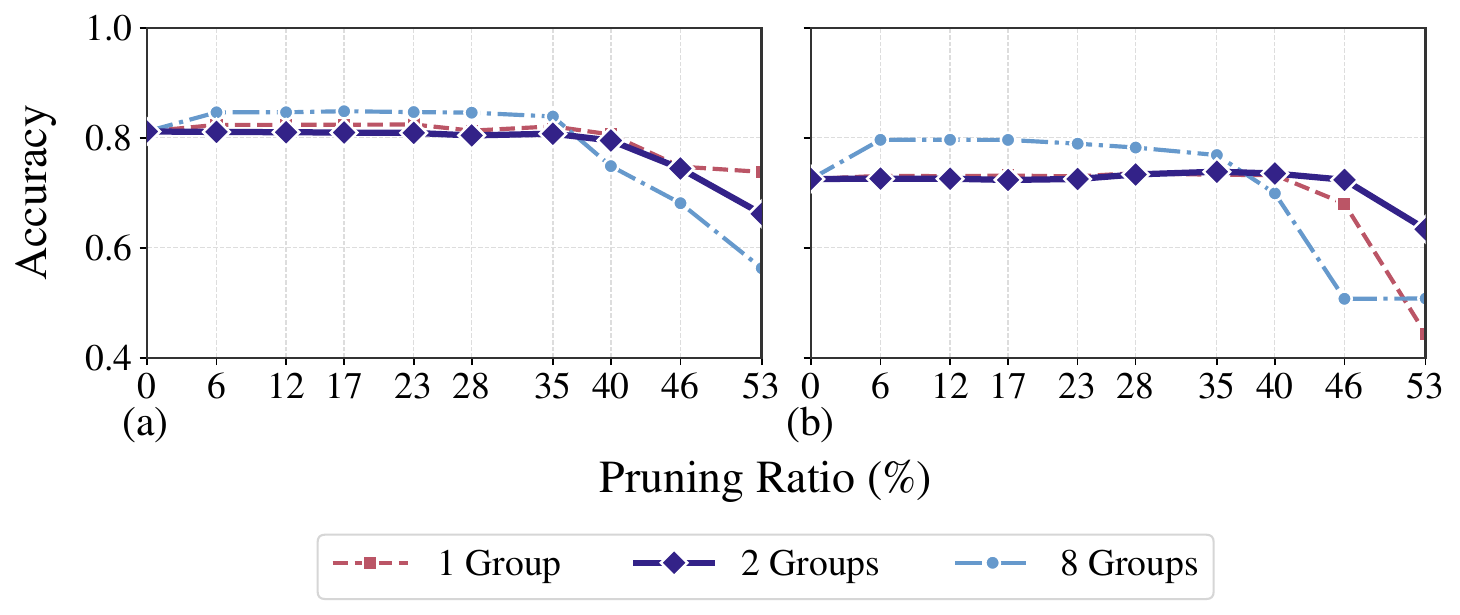}
    \vspace{-25pt}
    \caption{Accuracy of \method{} under different grouped-query attention configurations (1, 2, and 8 key-value groups) across pruning ratios on DaliaHAR under (a) full modalities and (b) 4-modality dropout. }

    \vspace{0pt}
    \label{fig:gqa_groups}
\end{figure}

We ablate \method{} from two perspectives. First, we test whether the importance scores learned by \gate{} can approximate a strong gradient-based saliency teacher. Second, we study how the number of groups in \attn{} affects robustness under joint pruning and modality missingness.

\smallskip
\noindent\textbf{Amortizing Taylor-based saliency with \gate{}.}
We compare the MAESTRO-based instantiation of \method{} on DaliaHAR against two references: random pruning and a Taylor-based saliency teacher. Taylor saliency provides strong first-order supervision during training, but it is impractical at deployment because it requires gradient computation and re-estimation when the input condition or modality availability changes. In contrast, \gate{} amortizes this process into a lightweight forward-pass importance predictor.

Figure~\ref{fig:bound} shows that \method{} remains consistently close to the Taylor-based teacher while staying well above random pruning across pruning ratios and modality-dropout settings. Under full modalities, \method{} achieves 0.80/0.78 accuracy at 28\%/40\% pruning, compared with 0.81/0.79 for the Taylor teacher and 0.54/0.45 for random pruning. Under 4-modality dropout, \method{} reaches 0.72 at 28\% pruning and 0.71 at 46\%, compared with 0.73 and 0.72 for the teacher. Overall, \method{} recovers 95--98\% of teacher performance across the evaluated settings without requiring gradient-based saliency estimation at inference time, remaining close to a strong Taylor-based reference.

\smallskip
\noindent\textbf{Sensitivity to the number of GQA groups.}
We next vary the number of grouped-query attention groups in \attn{} to study the trade-off between parameter sharing and robustness. Specifically, we evaluate 1, 2, and 8 groups, where 1 group corresponds to the strongest sharing. Figure~\ref{fig:gqa_groups} shows that the difference is small at light pruning but becomes more visible as pruning and missingness intensify. Under full modalities at 53\% pruning, the 1-group configuration achieves 0.78 accuracy, compared with 0.75 for 8 groups. Under 4-modality dropout, the gap becomes larger: at 28\% pruning, the 1-group configuration reaches 0.75 versus 0.70 for 8 groups, and at 46\% pruning, 0.65 versus 0.52.

Across conditions, the 2-group configuration remains consistently competitive and provides the most balanced trade-off between stronger parameter sharing and representational flexibility. We therefore use 2 groups as the default setting in the remainder of the paper.


\section{Conclusion}
We propose \method{}, a robust learning framework to enhance the deployability of multimodal models on edge devices. We posit that modality-aware pruning is necessary, especially when the observability of all modalities during inference is not guaranteed. Another key constraint is that fine-tuning–based pruning cannot always be applied, as backpropagation may be infeasible on edge devices and labeled data during inference may be unavailable. To address these practical limitations, we propose two innovations: \gate{}, which enables modality-aware zero-shot pruning trained via a saliency objective, and \attn{}, an efficient drop-in replacement for dense attention, a common performance bottleneck in multimodal models. Extensive evaluation across three applications, five baselines, and four compute platforms, from GPUs to mobile phones, demonstrates \method{}'s ability to maintain accuracy while substantially reducing parameters under extreme missingness.
\newpage

\bibliographystyle{ACM-Reference-Format}
\bibliography{ref}
\end{document}